# Fuzzy neural networks to create an expert system for detecting attacks by SQL Injection.

Lucas Oliveira Batista[1], Gabriel Adriano de Silva[1], Vanessa Souza Araújo[1], Vinícius Jonathan Silva Araújo[1], Thiago Silva Rezende[1], Augusto Junio Guimarães[1], Paulo Vitor de Campos Souza[1,2]

*(1) Faculdade Una de Betim, Email: lobatista@outlook.com.br, adriano.gabriel7@gmail.com, v.souzaaraujo@yahoo.com.br, vinicius.j.s.a22@hotmail.com, silvarezendethiago@hotmail.com, augustojunioguimaraes@gmail.com*

*(2) Centro Federal de Educação Tecnológica de Minas Gerais-CEFET-MG, Email: goldenpaul@informatica.esp.ufmg.br*

**Abstract:** Its constant technological evolution characterizes the contemporary world, and every day the processes, once manual, become computerized. Data are stored in the cyberspace, and as a consequence, one must increase the concern with the security of this environment. Cyber-attacks are represented by a growing worldwide scale and are characterized as one of the significant challenges of the century. This article aims to propose a computational system based on intelligent hybrid models, which through fuzzy rules allows the construction of expert systems in cybernetic data attacks, focusing on the SQL Injection attack. The tests were performed with real bases of SQL Injection attacks on government computers, using fuzzy neural networks. According to the results obtained, the feasibility of constructing a system based on fuzzy rules, with the classification accuracy of cybernetic invasions within the margin of the standard deviation (compared to the state-of-the-art model in solving this type of problem) is real. The model helps countries prepare to protect their data networks and information systems, as well as create opportunities for expert systems to automate the identification of attacks in cyberspace.

## I. Introduction

The technological race is one of the most discussed topics in the world today, generating great tools for commerce, industries, automation, communication, schools, military service, governments, among other vital sectors of the economy [1]. As a result, the search for high-speed information processing and transmission and high system performance tends to move in parallel with the security and integrity of the data, making information traffic dangerous and subject to constant attacks by hackers. Parallel to this technological growth, we also identify a significant evolution of so-called cyber-attacks, procedures that aim to compromise information security and computational systems [2].

With globalization and increasing societal dependence on software systems, information and data of paramount importance to companies and individuals worldwide travel instantly via the internet, catching the attention of cybercriminals, where they seek to invade systems or intercept information for use for their benefit, or harmful to the organizations they attack, making the impacts of such attacks more and more high [3].

This article proposes the use of hybrid models based on neural networks and fuzzy systems to build systems specialized in the cybernetic invasion, based on fuzzy rules. The attack that will be the target of this

In this case, the neural networks of the neural nets are similar to those proposed by Demertzis et al. [3]. The system introduced in the article will be able to generate rules based on the results obtained through tests, using nebulous logic neurons. To avoid over fitting and assisting in the definition of the network topology, training models based on the extreme learning machine and regularization theory will be used to find the most significant neurons in the cybernetic invasion problem.

The use of fuzzy neural networks has been used in several branches, such as economics [4], for the recognition of human faces in the 3D form [5], selection of characteristics [6], rainfall forecasting [7]. This intelligent technique is already used in the field of information security as a methodology to detect attacks in network traffic [8] and recognition of attack patterns [9]. The fuzzy neural network proposed in [10] seeks to perform binary classification tests to create the rules for the identification of expert systems in cybernetic attacks by SQL Injection.

The article is organized as follows: In section II we have the technical reference, with a definition of essential concepts involved in the development of the work. At section III are presented the description of the process of using the hybrid artificial intelligence model for identification and system specialists in cyber-attacks, with details of specific procedures and concepts used. Finally, in section IV the conclusions are presented.

# 2. Literature Review

### A. Cyberspace

The term cybernetic space was first employed in a novel written by Willian Gibson "Neuromancer" [13]; we can consider it as the metaphor that describes non-physical territory created by computational means, notably the internet, where individuals and corporations, alone or in groups, members of companies, public agencies or governments, can communicate, conduct research and traffic data in general, using Information and Communication Technologies (ICT) as support for its operation [11]. Actions in cyberspace are classified as offensive, exploitative or protective, and offensives can even impact national security [12].

### B. Cyber Attacks

In the world situation, cyberspace is an area in which, despite having an understanding of the need for security, there are no measures implemented in a systematic and articulated way that can guarantee the reliability and preservation of the systems used. In an early form, nations have been preparing to avoid or minimize cyber-attacks on networks and government information systems, as well as all other segments of society [11].

The attacks can happen in a physical way, where the devices containing the information are easily accessible, as well as modems, cables and physical storage media [16]. By human means, the attack is employed by social engineering, and by the logical medium, in which techniques like the invasion to overthrow services (DDoS) are used where the attacker works to overload the system in focus.

Techniques that exploit the vulnerability of access ports, shipping mass virus and malware, or even decoders of passwords, the last one that consists of a script that tries to decipher passwords [16].

Another essential aspect to be considered relates to cybercrime, mainly due to the harmful effects that may result from the misuse of information and communication systems by malicious persons. Despite the efforts of some sectors of the Public Administration, there are still loopholes in Brazilian law, and there are no laws for some types of actions that are already considered crimes in other countries. Also, there is no clear policy grounding a lawsuit against another state that has somehow affected critical national infrastructure by cybernetic means and many issues are still deserving some treatment, and this progress is being slower than the situation requires [12].

### C. SQL Injection

Structured Query Language, or just SQL, is the default language for interacting with relational databases. In it, we can do the main tasks related to data manipulation in database structures [20].
SQL Injection is a type of cyber-attack that takes advantage of faults in systems that generally have a database communication through SQL commands, and for that reason is considered a kind of attack is straightforward. In this invasion process, the attacker can insert a custom and undue SQL statement inside a query (SQL query) through the data entries of a program, such as forms or URL of an application. In the fields destined for user information, these commands are performed, i.e., SQL commands are displayed, however, because of this failure in the applications they end up causing changes in the database or inappropriate access to the application [21].

A cracker can get any stealth data held in the database of a server computer through SQL injection attacks, including depending on the version of the database, you can also enter malicious commands and get full permission to the machine where the bank is running [22]. Figure 1 shows essential steps for an attack by SQL Injection.

### D. Cyber Defense

According to Kshetri's definitions: "Cybernetic security and understood as the art of assuring the existence and continuity of the Information Society of a nation, guaranteeing and protecting, in the Cyber Space, its information assets and its critical infrastructures" [14]. We can also characterize it as a set of defensive, exploratory and offensive actions, in the context of planning, carried out in cyberspace, with the purpose of protecting our information systems [12].

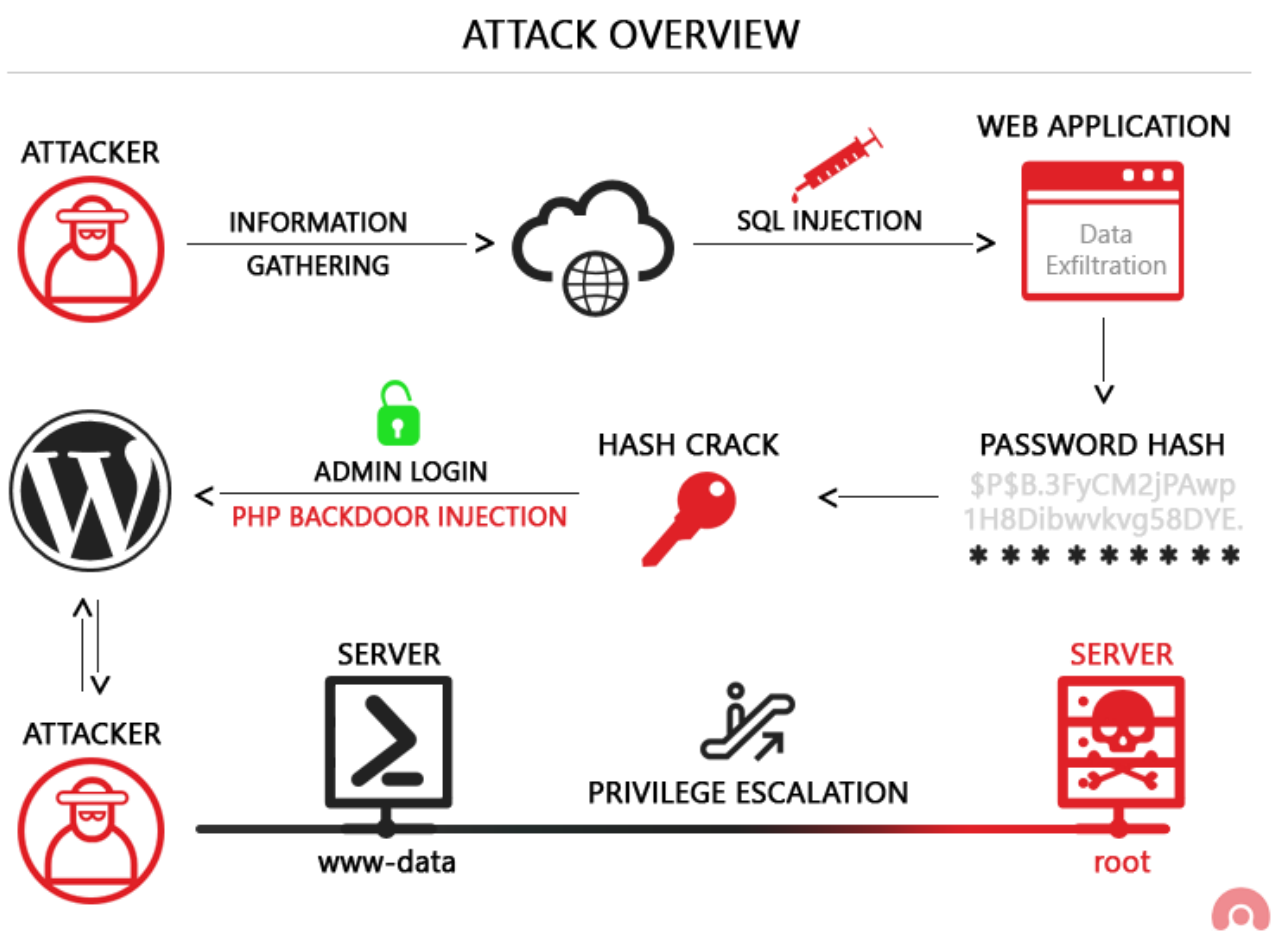

Figure 1. SQL Attack overview. Available in: [https://www.acunetix.com/blog/articles/exploiting-sql-injection-example/](https://www.acunetix.com/blog/articles/exploiting-sql-injection-example/)

Keeping data secure is a 21st-century challenge for Brazil. There is an increasing emphasis on cyber security as a strategic function of the state, and it is essential for the maintenance of the critical infrastructures of the country. It can be said that the country to develop cannot relinquish the security of its cyberspace. According to the provisions of the National Defense Strategy (END) [17], the responsibility of coordinating the actions of Cyber Defense is entrusted to the Army within the Armed Forces.

Due to this responsibility, the Brazilian Army is creating the Center for Cyber Defense, has been articulating in this area and consolidating its position [12]. It is therefore observed that there is a notion of the importance of taking actions that favor and allow the establishment of security in cyberspace, although the exact size of what this may represent is not known, as well as the measures that can and do be adopted to achieve this goal.,

### E. Artificial neural networks

Artificial neural networks are intelligent models that use in their structures the logical neuron, trying to simulate the processing of information of the human brain through a system of several interconnected artificial neurons that are united employing synaptic connections. In a simplified way, an artificial neural network can be seen as a graph where the nodes are the neurons, and the links function the synapses [18].

Artificial neural networks are differentiated by their architecture and the way the weights associated with the connections are adjusted during the learning process. Learning is the way in which the neural network captures the information provided by the inputs and through the relationships and the synaptic weights make decisions about the central theme of the database. The architecture of a neural network restricts the type of problem in which the network can be used, and is defined by the number of layers (single layer or multiple layers), number of nodes in each layer, type of connection between nodes feed forward or feedback) and by their way of acting [18]. Figure 2 shows the structure of a neural network with multiple layers. It also highlights the neurons and their synaptic connections.

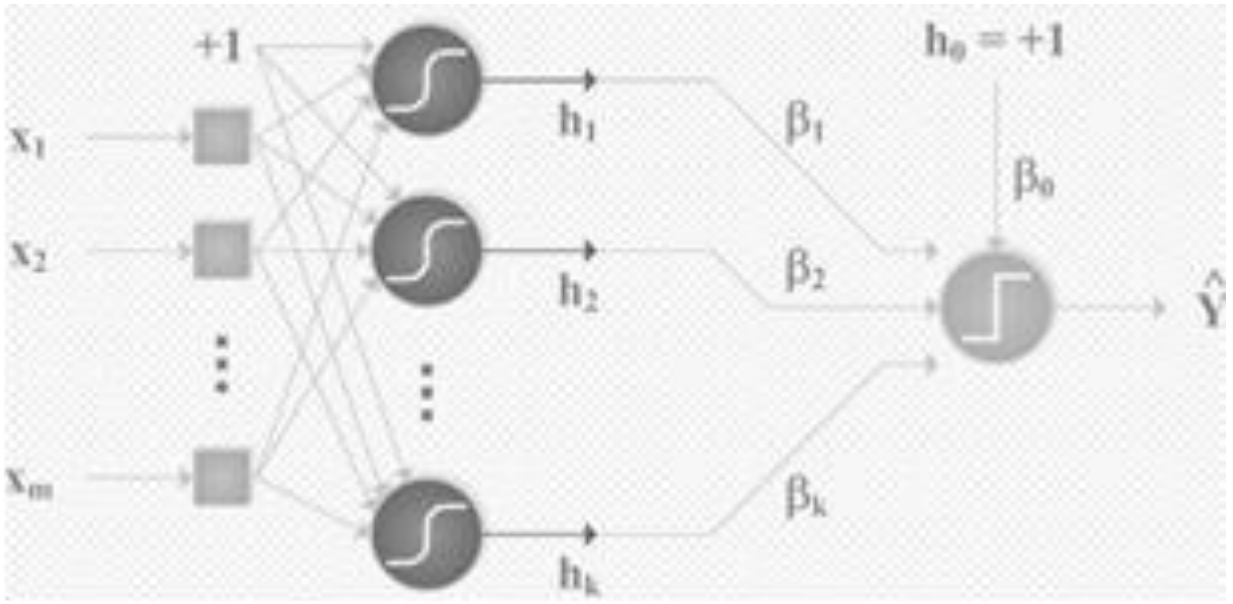

Figure 2. Example of an artificial neural network with multiple layers and only SQL Injection Attacks [35].

### F. Fuzzy Systems

The use of fuzzy systems is necessary in cases where the classical logic approach becomes unfeasible for solving a problem due to the nature of its complexity [19]. The best-known methods are subject to sudden changes to solve problems due to the simplification of the real model, but the fuzzy systems have resources (pertinence functions, rules, and aggregation operators) that allow a more accurate approximation to the actual model, avoiding that the solution generated by the fuzzy system differs considerably from the real model. Figure 3 presents the main elements of fuzzy logic: its inputs, the process of transforming input data into fuzzy elements, the creation of input fuzzy sets, the set of rules and inferences, the obtaining of fuzzy response sets, the defuzzification that is to make the values obtained according to the inputs of the system and the SQL Injection Attacks in an expected way.

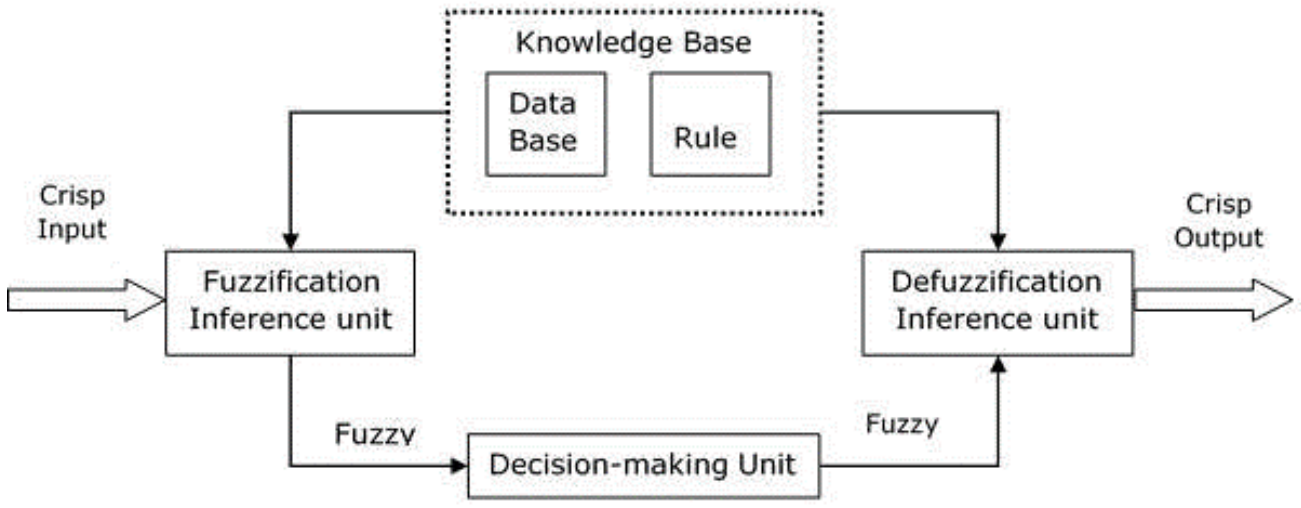

Figure 3. Concepts present in fuzzy logic. Available in: https://www.tutorialspoint.com/fuzzy_logic/fuzzy_logic_inference_system.htm

### G. Fuzzy neural networks - General Concepts

Fuzzy neural networks use the structure of an artificial neural network (ANN), where classics of artificial neurons are replaced by fuzzy neurons [7] [9]. These neurons are implemented using triangular rules that generalize the operations of union and intersection of classical sets allowing them to be used in fuzzy set theory. Thus, the neural network is now seen as a linguistic system, preserving the learning capacity of RNA [8].

They provide a network as the topology and allow the use of a wide variety of learning processes with databases from various contexts. The main feature of these networks is their transparency, allowing the use of a priori information to define the initial network topology and allowing the extraction of valuable information from the resulting topology after training in the form of a fuzzy set of rules [10].

The parameter update, types of neurons used and the training algorithm differ concerning the created network topologies. There are fuzzy neural networks that use extreme learning machine [26] to update internal parameters of the RNN where their logical neurons are unineurons [24], nullneurons [25], and andneurons [10].

Fuzzy neural network models are used in diverse contexts of economics and science in recent years. The techniques used to create specialized systems act in settings such as sleep ECG signal detectors [36], Architectural Foundation Selection [37], control of the wastewater treatment process with multiobjective operation [38]. Linear regression models have also studied in fuzzy neural networks in [39], and in [40] the hybrid model is used for software effort estimation. Therefore it is notorious the scope of areas and approaches that these models can act.

# 3. SQL Injection Attack Identification Methodology (Model-bioHAIFCS)

The paper on network anomaly detection based on neural network evolution written by Konstantinos Demertzis and Lazaros Iliadis [3], describes an intelligent system of machine learning, where part of the system works looking for known threats, and another part tries to detect probable threats according to abnormal activities that take place in the order. The detection system is simple, it generates a state being treated as usual, and all the signals outside the edge of that state are processed as an anomaly, so the detection algorithm learns continuously while the system is active in the network, is more and more need.

The methodology used in the article was the Spiking Artificial Neural Networks (SANN) [3], which uses an Evolving Connectionist System (eCOS) and Multi-Layer Feed Forward ANN approach to classify the exact type of intrusion or network abnormality with minimal computational potential. SANN is a set of modular systems based on node connections. The system organizes itself continuously, in line mode, adapting itself from the input data, being able to function or not in a supervised way. The SANN is also being applied to several other complex real-world problems, proving to be quite capable. The name of the developed model is called a bio HAIFCS (Hybrid Evolving Spiking Anomaly Detection Model), which works on the peaks that occur in the system, while the neurons are used to monitor the algorithm using One Pass learning.

Traffic-oriented data is used by importing the classes, which use the variable Population Encoding (control variable from data conversion of the sample to the actual value in the time peaks). Data were classified into two types, Class 0, is the class of standard results. Class 1, corresponding to abnormal results. When there is verification, and the result is 0, the eSNN classification process is repeated, but with appropriate data vectors. If the result continues 0, the process is terminated. When the result is Class 1, a neural network of two layers is used to recognize the pattern of the type of attack, using all the resources of the KDD cup database [40], if it happens in the hidden layer, 33 neurons are used. The results of the process are presented to the network administrator as an alert, and the HESADM graphical model can be analyzed in figure 4 [3].

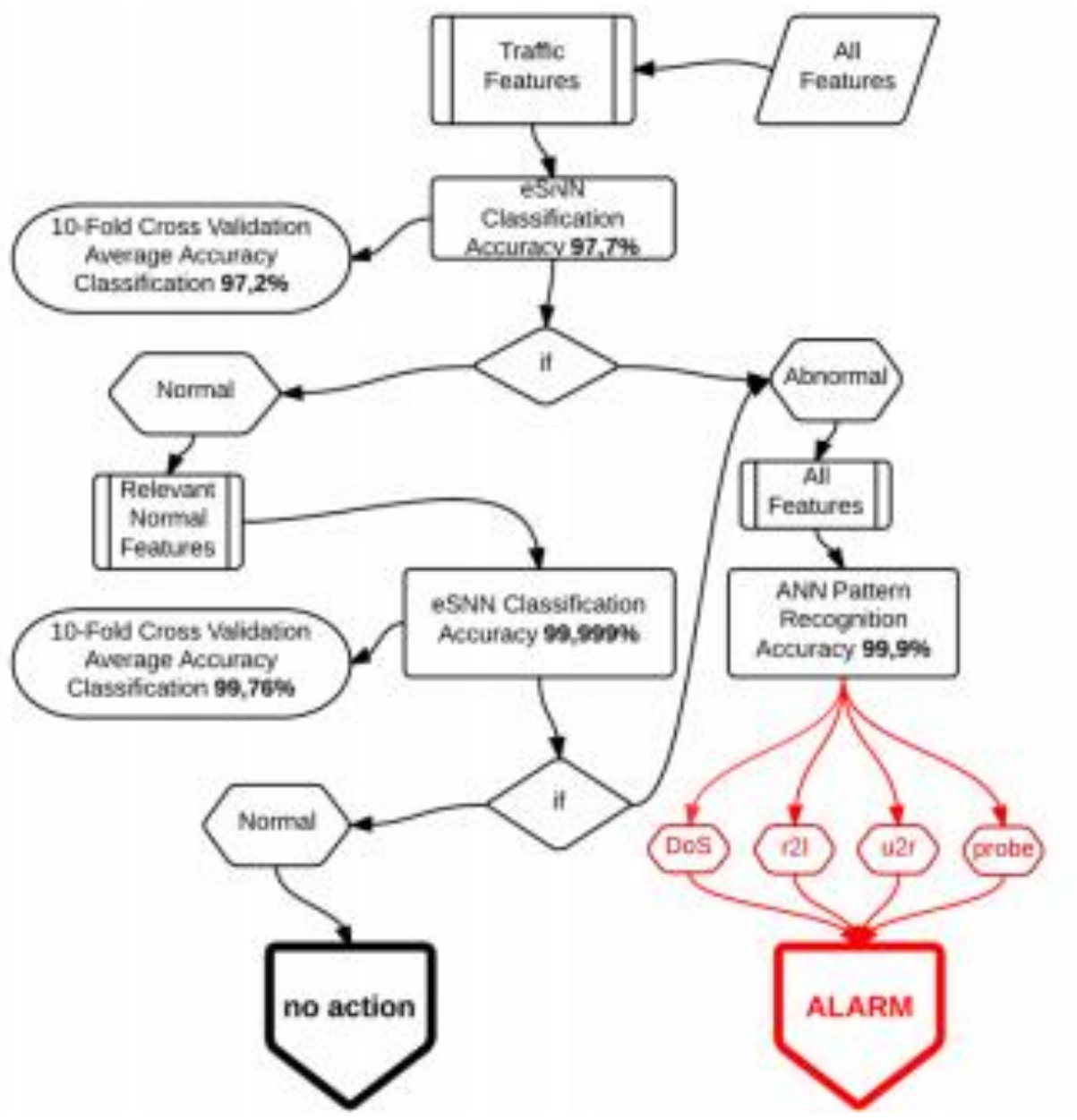

Figure 4- HESADM model [3].

The dataset used for the tests were from the KDD Cup 1999 [40], created in the Lincoln Lab of MIT [23], a set that according to the authors of HESADM is the most popular used. The collection contains data simulating a US Air Force network. "The event analysis method includes a connection between a source IP address and a destination IP, during which a sequence of TCP packets are exchanged, using a specific protocol and a strictly defined operating time. The database used includes a list of 13884 SQL statements that have been selected by various sources. Twelve thousand eight hundred eighty-one of them are malicious (SQL Injections), and 1003 are

legitimate — the correlation of SQL statements with the type of SQL injections. Finally, the n-gram technique was used to investigate the relationship of the SQL statement sequence with the syntax of the SQL injection commands [3].

In figure 5 the excellent performance and reliability of the scheme proposed in [3] are shown. It presents the results of the categorization with the same SQL Injection dataset and employing Cross-Validation with ten k-fold, and other Machine Learning approaches. [3] The model reached the result of 99.6%.

| SQLI Dataset | |
|---|---|
| **Classifier** | **Accuracy** |
| **MFF ANN with GA** | 99.6% |
| **RBFNetwork** | 97.3% |
| **NaiveBayes** | 95.6% |
| **BayesNet** | 98.7% |
| **SVM** | 98.5% |
| **k-NN** | 98.3% |
| **Random Forest** | 99.1% |

Figure 5- Comparison of various approaches for the SQLI dataset

# 4. Fuzzy Neural Network for SQL-Injection Attack Detection

The model for detection of cyber-attacks was initially proposed for classification of binary patterns. The model presented is a grouping of concepts of the models introduced in [10], [24] and [25]. Figure 6 shows the architecture of the fuzzy neural network used in the detection of SQL-Injection attacks.

The first two layers of the model are considered a fuzzy inference system, capable of extracting knowledge from the data and transforming them into fuzzy rules. These rules aid the construction of automated systems for detecting SQL Injection attacks according to the characteristics simulated by American computers. Differently addressed in [24] the third layer is composed of a simple neuron that has a function of activation the methodology proposed by [27], called leaky ReLU.

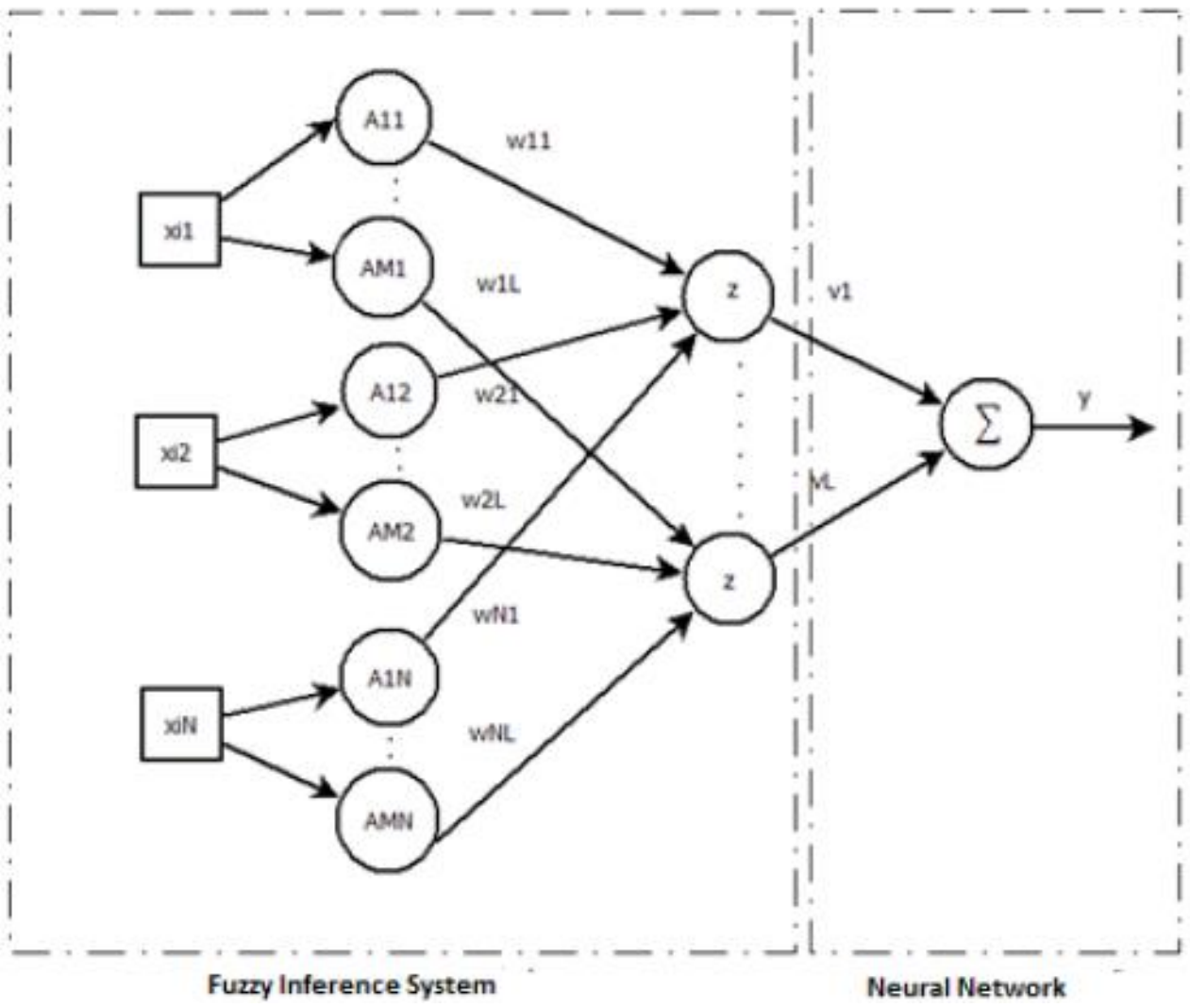

Figure 6- Fuzzy neural network model use in this paper [24].

The first layer is composed of fuzzy neurons whose activation functions are Gaussian membership functions of the fuzzy sets defined according to the partition of the input variables. For each input, variable $x_{ij}$ are defined $M$ fuzzy sets $A^m$, $m$ of 1,... $M$ whose membership functions are the activation functions of the corresponding neurons. Therefore, the SQL Injection Attacks of the first are the degrees of pertinence associated with the input values, that is, $a_j = \mu^A{}_m$ for $j = 1, .., N$ and $m = 1, ..., M$, where $N$ is the number of inputs and $M$ is the number of fuzzy sets for each input variable [10, 24, 25]. For the neurons of the first layer, the values of bias and synaptic weights are defined at random in the interval of [0, 1]. In this work, we will use the total combinations of fuzzy sets generated for each variable when $N$ is less than or equal to 6 [24]. When $N$ has high numeric values, we perform the random selection of a membership function for each input variable, where $M$, in this case, will be twice the value of input space samples, limited to 500 membership functions. Then we use the SQL Injection Attacks of the fuzzy neurons of the model to define many candidate neurons ($L_c$) that represent a percentage of $L$ where $L_c < L$. By definition, when $L$ <200 is used $L_c$ = 100% of $L$, case otherwise the chosen rate can select the candidate neurons. This percentage allows the choice of the most essential neurons of the first layer [25].

The second layer is composed of $L_c$ fuzzy logic neurons, where we highlight the unineuron proposed by [28]. Each neuron performs a weighted aggregation of some outlets (and not all of them due to the neuron selection technique) of the first layer along with bias and randomly defined weights of the unineurons [24]. Logical neurons are functional units that combine logical aspects of processing with learning ability through the system of fuzzy rules. They can be seen as multivariable nonlinear transformations between unit hypercube, $[0, 1] \rightarrow [0,1]^n$ [19]. Thus, neurons *and* and *or* (figure 7) add the values of fuzzy relevance **a** = [$a_1$, $a_2$, ... , $a_3$] initially combining them individually with their weights **w** = [$w_1$, $w_2$, ... , $w_3$], **a, w** ∊ $[0,1]^n$ to combine these results in the following way [19]:

$$z = or(a, w) = S_{i=1}^{N}(a_i \; s \; w_i) \tag{1}$$

$$z = and(a, w) = T_{i=1}^{N}(a_i \; s \; w_i) \tag{2}$$

where $T$ and $t$ are *t-norms* (product) and $S$ and $s$ are *s-norms* (probabilistic sum).

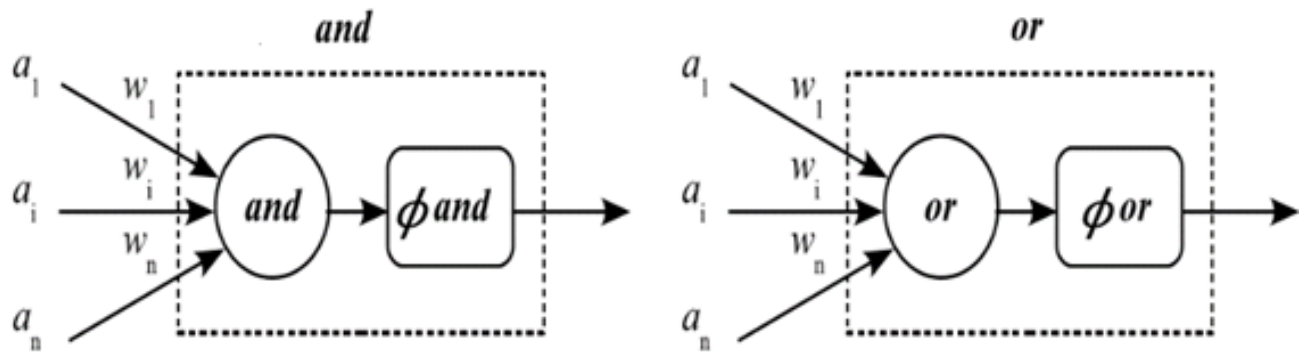

Figure 7. Schematic of neurons *and* and *or [28]*

Uninorms are the generalization of t-norms and s-norms by relaxing the constraints related to the neutral elements. Instead of values 0 and 1 for t-norm and s-norm respectively, the neutral element is allowed to assume values in the unit interval. One of the main characteristics of the uninorm is that it no longer has the so-called neutral element, now being called the entity element [28]. Through this identity element the uninorms extend *t-norms* and *s-norms* by varying the value *g* in the interval between 0 and 1 allowing the alternation between an s-norm (*g* = 0) and t-norm *(g* = 1) [28]. The uninorm used in this work is expressed as follows

:

$$U(x,y) = \begin{cases} gT(\frac{x}{g}, \frac{y}{g}), if..x, y \in [0, g] \\ g + (1-g)S(\frac{x-g}{1-g}, \frac{y-g}{1-g}), if...x, y \in [g,1] \\ \max(x,y) or..\min(x,y), otherwise \end{cases} \tag{3}$$

The unineuron uses the concepts of uninorm to perform more simplified operations according to the functions of activation of the nebulous neurons. Its formatting allows the unineuron to use either concepts of a neuron *and*, or a neuron *or*. In [28] explain important concepts about a *unineuron*. The processing of neurons occurs at two levels. At the first level of $L_1$ locations the input signals are combined individually with the weights. In the second, at global level $L_2$, a global aggregation operation is performed on the results of all first-level combinations. Traditional logical neurons use t-norms and s-norms to perform the described operations.

Given a collection of pairs Ω ($a_i$, $w_i$), with $a_i$ ∈ [0, 1] it's an entrance and $w_i$ ∈ [0, 1] the following steps should be followed to perform the weighted aggregation [28]:

- Transform each pair ($a_i$, $w_i$) in a single value $b_i = h(a_i, w_i)$;
- Calculate the unified aggregation of transformed values **U** ($b_1$, $b_2$, ..., $b_m$), where m is the number of inputs.

The function *h* is responsible for transforming the inputs and corresponding weights into individual transformed values [28]. A formulation for the *h* function can be visualized:

$$h(w, a) = w * a + w * g \tag{4}$$

Using the weighted aggregation reported in (4) we can write the unineuron that can be visualized:

$$z = uni(x, w, g) = U_{i=1}^{N}(x_i, w_i) \tag{5}$$

Fuzzy rules can be extracted through the network topology in conjunction with unineuron logical neurons. The following is an example of combinations of rules:

**Rule1**: **If** $x_{i1}$ is $A^1_1$ with certainly $w_{11}$...

**and/or** $x_{i2}$ is $A^1_2$ with certainly $w_{21}$...

**then** $y_1$ *is* $v_1$

**Rule2**: **If** $x_{i1}$ is $A^2_1$ with certainly $w_{12}$... (6)

**and/or** $x_{i2}$ is $A^2_2$ with certainly $w_{22}$...

**then** $y_2$ is $v_2$

**Rule3**: **If** $x_{i1}$ is $A^3_1$ with certainly $w_{13}$…

**then** $y_3$ is $v_3$

**Rule4**: **If** $x_{i2}$ is $A^2_2$ with certainly $w_{23}$…

**then** $y_4$ is $v_4$

After defining the candidate neurons, the final architecture of the network is determined using the selection of a subset of these neurons. When performing this procedure, we are implementing an optimum subset of values, which can be viewed as a variable selection problem, returning the most significant neurons ($L_s$) based on a cost function [24]. Analogously, we can interpret this selection as the choice of the best set of rules capable of representing the input space. The learning algorithm assumes that the SQL Injection Attack of the second layer of the fuzzy neural network composed of all the most significant neurons ($L_s$) can be written as

$$f(x_i) = \sum_{l=0}^{ls} v_l z_l(x_i) = z(x_i)v \quad (7)$$

Where **v** = $[v_0, v_1, v_2, ..., v_{Ls}]$ is the weight vector of the SQL Injection Attack layer and **z** ($\mathbf{x}_i$) = $[z_0, z_1(x_i), z_2(x_i) ..., z_{Ls}(x_i)]$ is the argument vector (line) of the second layer to $z_0 = 1$. In this context, **z** ($\mathbf{x}_i$) is considered such as the non-linear mapping of the entry for an $L_s + 1$ dimensional fuzzy features, performed using the neurons selected. As the weights that bind to the two first layers are assigned random shape and the only parameters were the weights of the SQL Injection Attack we can see equation (7) as a simple linear regression model allowing the problem of the choice of best subsets of neurons that will be can be idealized as a model of linear regression for selection problems [29]. A very used to perform model selection was created by [30] and is known as Algorithm Least Angle Regression (LARS). LARS is a regression algorithm for data with that it is not able to estimate only the regression coefficients, but also a subset of regressors candidates to be included in the final model. To we evaluate a set of K different samples (xi, yi), where $\mathbf{x}_i = [x_{i1}, x_{i2}, ..., x_{iN}] \in \mathbb{R}$ and $yi \in \mathbb{R}$ for $i = 1, ..., K$, the cost function of this algorithm of regression can be defined as:

$$\sum_{i=1}^{K} \| \; z(x_i)v - y_i \; \|_2 + \lambda \| \; v \; \|_1 \quad (8)$$

Where $\lambda$ is a regularization parameter estimated using the cross validation. The first term of (8) corresponds residual sum of squares (RSS). That term decreases as the error of training also fall. The second term is a regularization term $l_1$. This expression is used because it improves the generalization of the network avoiding the over-adjustment [31] and can generate sparse models [32]. Rewriting equation (8) we understand why LARS be used as a feature selection algorithm is:

$$\min_v RSS(v) st \| \; v \; \|_1 \leq \beta \quad (9)$$

Where $\beta$ is an upper bound on the $l1$-norm of the weights, a small value of β corresponds to a large amount of $\lambda$ and vice versa. This equation is also known as lasso [33]. When we use the lasso regression (also called $l1$-norm) to normalize models, we find that the method leads to results with spatial solutions, generating result vectors with many zeros, which represent data of no importance for the analyzed variables. Best selection of models [34]. The LARS algorithm can be used to perform the model selection, since, for a given value of $\lambda$, only a fraction (or none) of the regressors have equal weights other than zero. If $\lambda = 0$, the problem becomes unrestricted regression, and all weights are nonzero. As $\lambda_{max}$ increases from 0 to a given value $\lambda_{max}$, the number of nonzero weights decreases from $N$ to 0. For the problem considered in this work, the regressors $\mathbf{z_{ls}}$ are the SQL Injection Attacks of the significant neurons. Thus, the LARS algorithm can be used to select an optimal subset of the vital neurons that minimize (9) for a given value of $\lambda$, obtained through cross-validation. Using the concept of bootstrap and performing the intersection between supports, Bach [34] developed a model adjustment estimator, without the conditions of consistencies required by the lasso method. To this new procedure, he gave the name of Bolasso (bootstrap-enhanced least absolute shrinkage operator). This framework can be seen as a voting scheme applied to support of the lasso method. At the However, Bolasso can be seen as a consensus-building regime where it is maintained the most significant subset of variables on which all regressors agree when the aspect is the selection of variables [34]. The regressors to be included in the final model are defined according to the frequency with which each of them is selected through different tests. A consensus threshold is determined, say $\rho$ = 50%, and a regressor is included if chosen in at least 50% of the trials. In this article, the lasso bootstrap is used to define the network topology and select the most significant neurons. The extreme learning machine concepts [26] are applied to calculate the weights of the SQL Injection Attack layer and the neural aggregation network, present in the third layer of the model, performs the classification of cyber-attack patterns according to equation (10):

$$y = sign(f_{leaky\,\mathrm{Re}\,LU}(\sum_{j=0}^{ls} z_j v_j)) \qquad (10)$$

Where $z_0$ = 1, $v_0$ is the bias, and $z_j$ and $v_j$, j = 1, ..., $l_s$ are the SQL Injection Attack of each fuzzy neuron of the second layer and their corresponding weight, respectively. The leaky ReLU function is expressed by [27]:

$$f_{leaky\,\mathrm{Re}\,LU}(z,\alpha) = \max(\alpha z, z) \qquad (11)$$

This function of activation is now employed in problems of diverse natures, especially those where it demands a higher sensitivity in the results obtained by the fuzzy neural networks.

Finally, after the definition of the network topology, we calculate the vector of weights of the SQL Injection Attack layer $\mathbf{v} = [v_0, v_1, v_2, v_{Ls}]^T$. In this paper $\mathbf{v}$ is calculated using the pseudo Inverse of Moore-Penrose:

$$v = Z^{+} y \qquad (12)$$

where $Z^{+}$ is pseudo-inverse of Moore-Penrose of Z which is the minimum norm of the solution of the least squares for the weights of the exit. Z can be defined as:

$$Z = \begin{bmatrix} z_0 & z_1(x_1) & \ldots & z_{ls}(x_1) \\ z_0 & z_1(x_2) & \ldots & z_{ls}(x_2) \\ . & . & \ldots & . \\ z_0 & z_1(x_n) & \ldots & z_{l_s}(x_n) \end{bmatrix} \qquad (13)$$

The learning process can be summarized as shown below. It has four parameters:

• The number of fuzzy sets that will partition the input space, $M$.

• The percentage of candidate neurons, $L_c$.

• The number of bootstrap replications, $b$.

• The consensus threshold, $\rho$.

# 5. SQL Injection Detection Tests

To perform the SQL Injection detection tests and to enable the construction of an expert system based on the nature of the data, the KDD Cup 1999 database was used [40], which includes 13.869 cases of which 12.881 are malicious, and 988 are legitimate (0.0723%) of SQL Injection attacks. Unbalanced data sets are a particular case for classification problems where class distribution is not uniform across classes. Usually, they are composed of two categories: the majority (negative) and the minority (positive). From these characteristics, the parameters length, entropy, level of malice, confidence level, confidence, and Class were used. The expert system is based on the IF/THEN. The fuzzy neural network models (UNI-RNN [24] is a fuzzy neural network composed of unineurons (4), and AND-RNN [10] is composed of andneurons (2). They were compared with other classifier algorithms for the database: SVM (Support Vector Machine) [41], MLP (Multilayer Perceptron) [42], NB (Naive Bayes) [43] and C4.5 [44].

The test conditions were similar to proceeding in work done in [3], where the configurations and the use of the weka tool [45] were the same.

It should be noted that to avoid trends in the tests performed, all available samples were exchanged and 30 measurements of accuracy were collected from each of the bases evaluated in each model analyzed. The variables involved in the process were normalized with mean zero and variance one. All SQL Injection Attacks of the model were normalized to the interval [-1, 1].

For the fuzzy neural network model the optimal parameters *M*, *b* and *ρ* were found through cross-validation (70% training, 30% for test) using 10-k-fold. The ranges were as follows: *M* = {2, 3, 4}, *b* = {8, 16, 32}, *ρ* = {50%, 60%, 70%}. The value of Lc was arbitrated at 200 as in [24].

The results (Table 1) were provided with the tests performed on a desktop machine with Intel Core i5-3470 3.20GHz processor and 4.00 GB Memory.

Table I- Results of the tests of identification of attacks SQL Injection

| Model | Acuracy | AUC | Sensitivity | Test Time |
|---|---|---|---|---|
| UNI | 98.44(0.15) | 98.00(0.01) | 98.96(24.23) | 586.14(10.12) |
| AND | 98,46(0.21) | 98.00(0.01) | 97.94(0.37) | 771.69(108.48) |
| SVM | 96.79(2.71) | 96.14(2.76) | 96.87(1.42) | 468.97(48.76) |
| MLP | 97.99(7.15) | 91.87(6.03) | 87.65(7.87) | 714.18(56.96) |
| NB | 95.14(2.14) | 62.09(1.34) | 79.65(5.15) | 543.12(33.02) |
| C4.5 | 92.18(3.43) | 96.54(11.76) | 84.36 (3.45) | 268.31(7.65) |

**A. Interpretability of the resulting model**

The system presented useful results on the use of fuzzy rules for the construction of systems. We can highlight the example obtained when using two membership functions allowing the parameters are classified as "low" and "high." Figure 8 shows the generated fuzzy inference system and Figure 9 represents the equally spaced membership functions created in one of the 30 simulations performed,

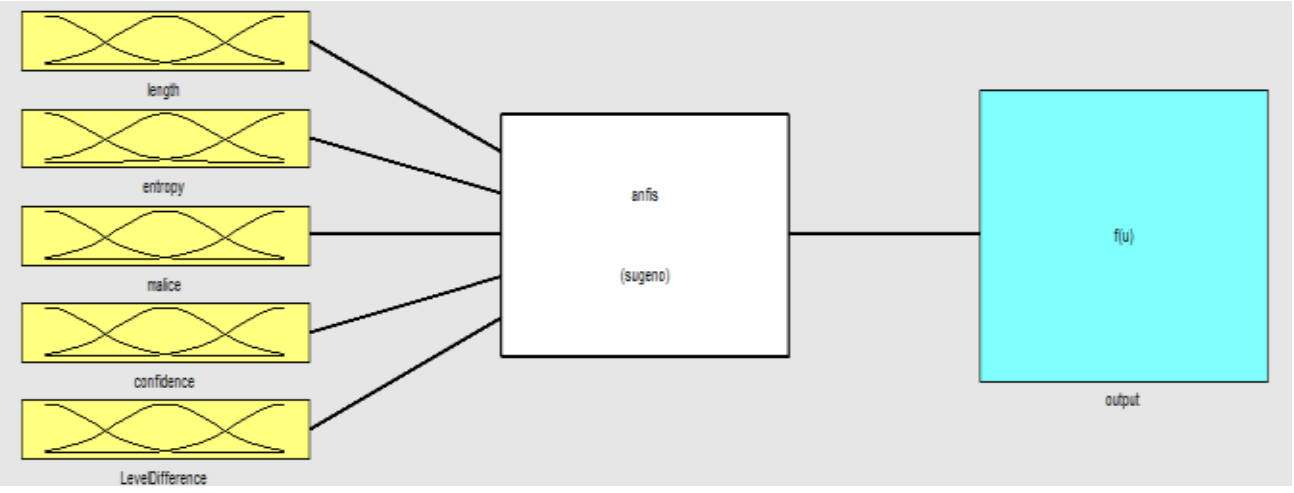

Figure 8- Fuzzy Inference System

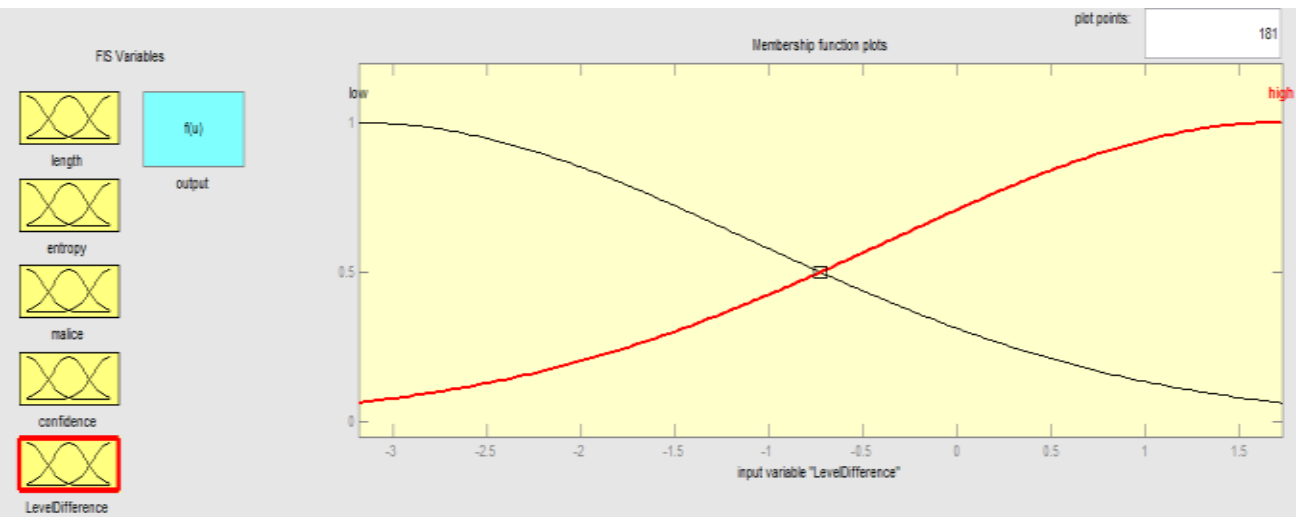

Figure 9- Gaussian Membership Functions

Fuzzy Rules Generated =

1. If (length is low) and (entropy is low) and (malice is low) and (confidence is low) and (level Difference is low) with certainly 10.3064 then (SQL Injection Attack is 0.1319)

2. If (length is low) and (entropy is low) and (malice is low) and (confidence is low) and (level Difference is high) with certainly 5.9251 then (SQL Injection Attack is 0.3012)

3. If (length is low) and (entropy is low) and (malice is low) and (confidence is high) and (level Difference is low) with certainly -12.4843 then (SQL Injection Attack is 0.4315)

4. If (length is low) and (entropy is low) and (malice is low) and (confidence is high) and (level Difference is high) with certainly -27.7771 then (SQL Injection Attack is 0.4718)

5. If (length is low) and (entropy is low) and (malice is high) and (confidence is low) and (level Difference is low) with certainly 13.3972 then (SQL Injection Attack is 0.4758)

6. If (length is low) and (entropy is low) and (malice is high) and (confidence is low) and (level Difference is high) with certainly -7.7441 then (SQL Injection Attack is 0.0071)

7. If (length is low) and (entropy is low) and (malice is high) and (confidence is high) and (level Difference is low) with certainly 1.8254 then (SQL Injection Attack is 0.6378)

8. If (length is low) and (entropy is low) and (malice is high) and (confidence is high) and (level Difference is high) with certainly -2.2822 then (SQL Injection Attack is 0.6378)

9. If (length is low) and (entropy is high) and (malice is low) and (confidence is low) and (level Difference is low) with certainly -3.6322 then (SQL Injection Attack is 0.1984)

10. If (length is low) and (entropy is high) and (malice is low) and (confidence is low) and (level Difference is high) with certainly 16.1158 then (SQL Injection Attack is 0.1310)

11. If (length is low) and (entropy is high) and (malice is low) and (confidence is high) and (level Difference is low) with certainly 1.2430 then (SQL Injection Attack is 0.1391)

12. If (length is low) and (entropy is high) and (malice is low) and (confidence is high) and (level Difference is high) with certainly -5.4759 then (SQL Injection Attack is 0.1392)

13. If (length is low) and (entropy is high) and (malice is high) and (confidence is low) and (level Difference is low) with certainly -10.4266 then (SQL Injection Attack is 0.1393)

14. If (length is low) and (entropy is high) and (malice is high) and (confidence is low) and (level Difference is high) with certainly 1.2170 then (SQL Injection Attack is 0.1319)

15. If (length is low) and (entropy is high) and (malice is high) and (confidence is high) and (level Difference is low) with certainly 4.6951 then (SQL Injection Attack is 0.3195)

16. If (length is low) and (entropy is high) and (malice is high) and (confidence is high) and (level Difference is high) with certainly 1.8021 then (SQL Injection Attack is 0.1316)

17. If (length is high) and (entropy is low) and (malice is low) and (confidence is low) and (level Difference is low) with certainly 13.7221 then (SQL Injection Attack is 0.1397)

18. If (length is high) and (entropy is low) and (malice is low) and (confidence is low) and (level Difference is high) with certainly -2.6547 then (SQL Injection Attack is 0.13198)

19. If (length is high) and (entropy is low) and (malice is low) and (confidence is high) and (level Difference is low) with certainly 8.1132 then (SQL Injection Attack is 0.13199)

20. If (length is high) and (entropy is low) and (malice is low) and (confidence is high) and (level Difference is high) with certainly -6.4444 then (SQL Injection Attack is 0.30120)

21. If (length is high) and (entropy is low) and (malice is high) and (confidence is low) and (level Difference is low) with certainly 1.7837 then (SQL Injection Attack is 0.30121)

22. If (length is high) and (entropy is low) and (malice is high) and (confidence is low) and (level Difference is high) with certainly 1.0121 then (SQL Injection Attack is 0.3012)

23. If (length is high) and (entropy is low) and (malice is high) and (confidence is high) and (level Difference is low) with certainly -4.8879 then (SQL Injection Attack is 0.3023)

24. If (length is high) and (entropy is low) and (malice is high) and (confidence is high) and (level Difference is high) with certainly -13.9030 then (SQL Injection Attack is 0.3124)

25. If (length is high) and (entropy is high) and (malice is low) and (confidence is low) and (level Difference is low) with certainly -5.8129 then (SQL Injection Attack is 0.3025)

26. If (length is high) and (entropy is high) and (malice is low) and (confidence is low) and (level Difference is high) with certainly -0.1328 then (SQL Injection Attack is 0.3026)

27. If (length is high) and (entropy is high) and (malice is low) and (confidence is high) and (level Difference is low) with certainly 3.5773 then (SQL Injection Attack is 0.2027)

28. If (length is high) and (entropy is high) and (malice is low) and (confidence is high) and (level Difference is high) with certainly 2.9249 then (SQL Injection Attack is 0.4028)

29. If (length is high) and (entropy is high) and (malice is high) and (confidence is low) and (level Difference is low) with certainly 10.7698 then (SQL Injection Attack is 0.3049)

30. If (length is high) and (entropy is high) and (malice is high) and (confidence is low) and (level Difference is high) with certainly -3.8301 then (SQL Injection Attack is 0.4331)

31. If (length is high) and (entropy is high) and (malice is high) and (confidence is high) and (level Difference is low) with certainly -12.5593 then (SQL Injection Attack is 0.43151)

32. If (length is high) and (entropy is high) and (malice is high) and (confidence is high) and (level Difference is high) with certainly 22.8950 then (SQL Injection Attack is 0.43152)

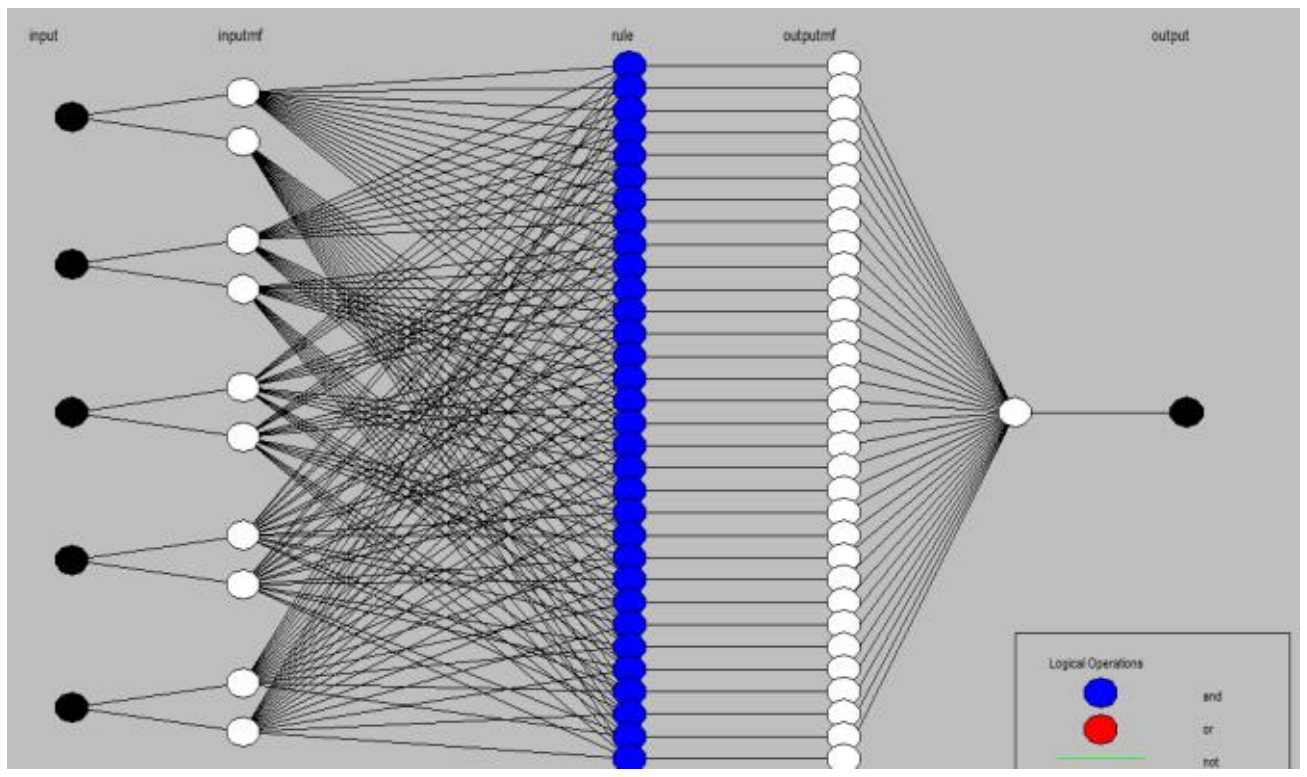

Figure 10- Result of the first layer of the model.

## 6. Conclusions

After the tests were performed using the fuzzy neural network model, we conclude that the results of the experiments prove that the creation of a specialized system helps to prevent cyber-attacks by encouraging the construction of intelligent applications.
Through the obtained results, it can be affirmed that the tests performed were satisfactory. Based on the results obtained by Demertzis [3], the model presented here is statistically equivalent to the HESADM model [3], reaching close to state of the art, (99% of HESADM, 98 of the FNN), and despite having been smaller than the comparative article [3], our model has a much greater interpretability by allowing specialists in the subject to validate the solutions of the system and that the fuzzy rules can create an object. Consequently, the results are also more interpretable and easy to use. Moreover, it is a work that has a great value in the scientific scope.

Further work may be done in the future exploring other configurations, fuzzy neural network models, and other tests may aid in improving the accuracy of the model.

## 7. Acknowledgments

The acknowledgments of this work are destined to the University Center UNA of Betim and the Federal Center of Technological Education of Minas Gerais - CEFET-MG.